\documentclass{bmvc2k}

\usepackage{tikz}
\usepackage{comment}
\usepackage{amsmath,amssymb} %
\usepackage{color}
\usepackage{multicol}
\usepackage{makecell}
\usepackage{multirow}
\usepackage{booktabs}
\usepackage{array}
\usepackage{pifont} 
\usepackage{wrapfig}

\title{XDGAN: Multi-Modal 3D Shape Generation in 2D Space} %

\addauthor{Hassan Abu Alhaija}{habualhaija@nvidia.com}{1}
\addauthor{Alara Dirik}{alara@pch-innovations.com}{2}
\addauthor{André Knörig}{andre@pch-innovations.com}{2}
\addauthor{Sanja Fidler}{sfidler@nvidia.com}{1,3}
\addauthor{Maria Shugrina}{mshugrina@nvidia.com}{1}

\addinstitution{
NVIDIA
}
\addinstitution{
PCH Innovations GmbH
}

\addinstitution{
University of Toronto
}

\runninghead{Alhaija \etal{}}{XDGAN - 3D Shape Generation in 2D Space}

\def\etal{\emph{et al}\bmvaOneDot}

\usepackage{amsfonts}

\makeatletter
\DeclareRobustCommand\onedot{\futurelet\@let@token\@onedot}
\def\@onedot{\ifx\@let@token.\else.\null\fi\xspace}

\def\etc{etc\onedot}

\def\etal{et~al\onedot}

\makeatother

\usepackage{xcolor}
\definecolor{darkred}{rgb}{0.7,0.2,0.1}
\definecolor{darkgreen}{rgb}{0,0.7,0}
\definecolor{orange}{RGB}{255,127,0}
\definecolor{ourpurple}{RGB}{127,127,204}
\definecolor{palgreen}{RGB}{51,179,179}
\definecolor{magenta}{RGB}{199,21,133}

\newcommand{\ourmodel}{XDGAN}

\begin{document}

\maketitle

\begin{abstract}
   
Generative models for 2D images has recently seen tremendous progress in quality, resolution and speed as a result of the efficiency of 2D convolutional architectures. However it is difficult to extend this progress into the 3D domain since most current 3D representations rely on custom network components. This paper addresses a central question: \textit{Is it possible to directly leverage 2D image generative models to generate 3D shapes instead?} To answer this, we propose XDGAN, an effective and fast method for applying 2D image GAN architectures to the generation of 3D object geometry combined with additional surface attributes, like color textures and normals. Specifically, we propose a novel method to convert 3D shapes into compact 1-channel geometry images and leverage StyleGAN3 and image-to-image translation networks to generate 3D objects in 2D space. 
The generated geometry images are quick to convert to 3D meshes, enabling real-time 3D object synthesis, visualization and interactive editing. 
Moreover, the use of standard 2D architectures can help bring more 2D advances into the 3D realm.
We show both quantitatively and qualitatively that our method is highly effective at various tasks such as 3D shape generation, single view reconstruction and shape manipulation, while being significantly faster and more flexible compared to recent 3D generative models.

\end{abstract}

\section{Introduction}\label{sec:intro}

Generative Adversarial Networks \cite{goodfellow2014gan} have achieved remarkable progress in generating high-resolution realistic images, typically using convolutional architectures such as \cite{BigGAN,karras2019style}. Extending these advances to the 3D domain remains a challenge and an active area of research, with newest methods introducing implicit or explicit 3D awareness into the GAN generative process \cite{niemeyer2021giraffe,chan2021pi,chan2021efficient,gu2021stylenerf, zhang2021ners}. In this work, we show that it is possible to directly leverage a 2D GAN architecture designed for images to generate high-quality 3D shapes. The key to our approach is parameterization of 3D shapes as 2D planar \emph{geometry images} \cite{Gu2002GeometryI}, which we use as the training dataset for an image-based generator.

Perhaps one of the greatest challenges to the development of generative models for 3D content is converging on the right representation. Direct extension of 2D pixel grids to 3D voxel grids \cite{Wu2016LearningAP} suffers from high memory demands of 3D convolutions, limiting resolution. Point cloud samples of surface geometry, while popular for generative tasks \cite{Achlioptas2018LearningRA,cai2020learning,Li2021SPGANS3}, are limited in their ability to model sharp features or high-resolution textures. Recently implicit 3D representations, such as NeRF \cite{mildenhall2020nerf} and DeepSDF \cite{Park2019DeepSDFLC}, as well as hybrids, have shown great promise for generative tasks \cite{Chen2019LearningIF,yan2022shapeformer,chan2021efficient}. However, these approaches typically are too slow for interactive applications and their output does not interface with existing 3D tools, calling for costly or lossy conversion of models before use. To this date, 3D meshes, augmented with normals and textures, remain the most widely adopted 3D representation in 3D software, movies and games. Our method uses 2D generative architectures to produce fixed-topology textured meshes with a single forward pass, and is thus immediately practical. 

\begin{figure}[t]
  \centering
  \subfloat{%
    \begin{tabular}{@{}cc@{}}
      \multicolumn{2}{c}{\textbf{Generation}}                                                          \\
      \includegraphics[width=0.22\textwidth]{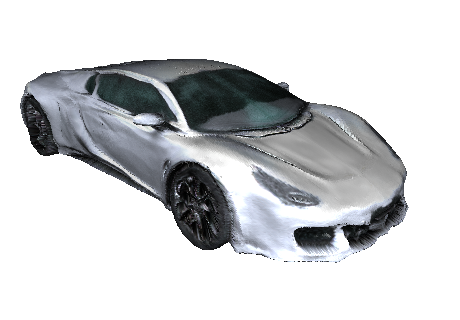} &
      \includegraphics[width=0.22\textwidth]{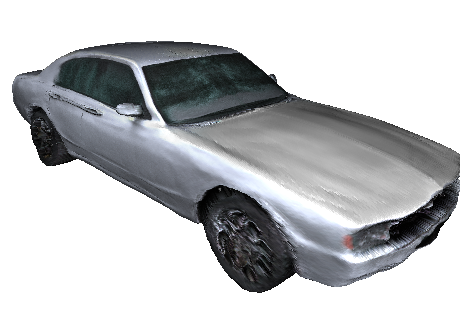}                               \\
      sample 1                                                                              & sample 2 \\
      \includegraphics[width=0.22\textwidth]{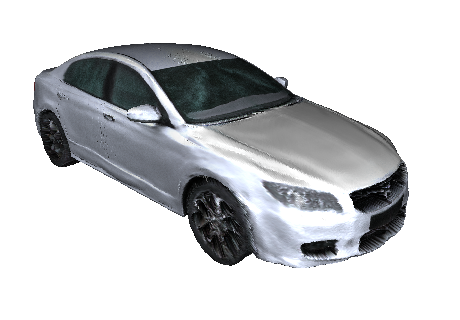}                    &
      \includegraphics[width=0.22\textwidth]{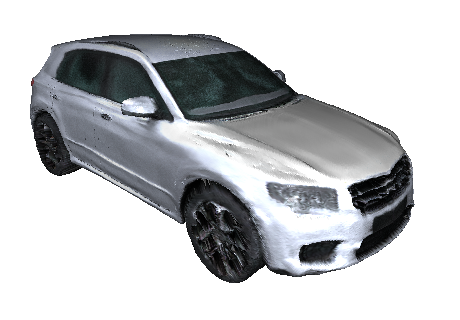}                               \\
      sample 3                                                                              & sample 4 \\
    \end{tabular}
  }
  \hspace{0.1em}
  \vline 
  \hspace{0.1em}
  \subfloat{%
    \begin{tabular}{@{}cc@{}}
      \multicolumn{2}{c}{\textbf{Projection}}                                                     \\
      \includegraphics[width=0.22\textwidth]{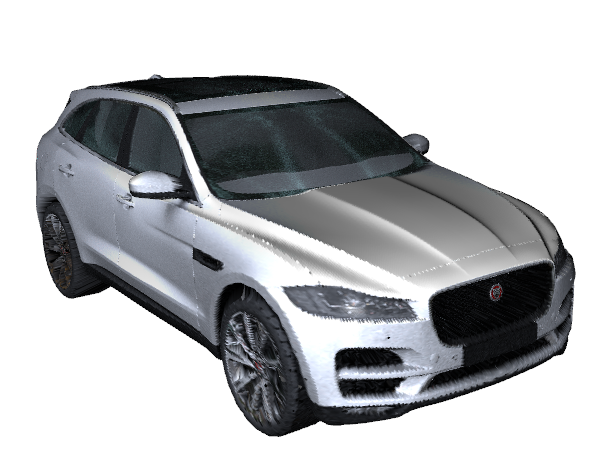} &
      \includegraphics[width=0.22\textwidth]{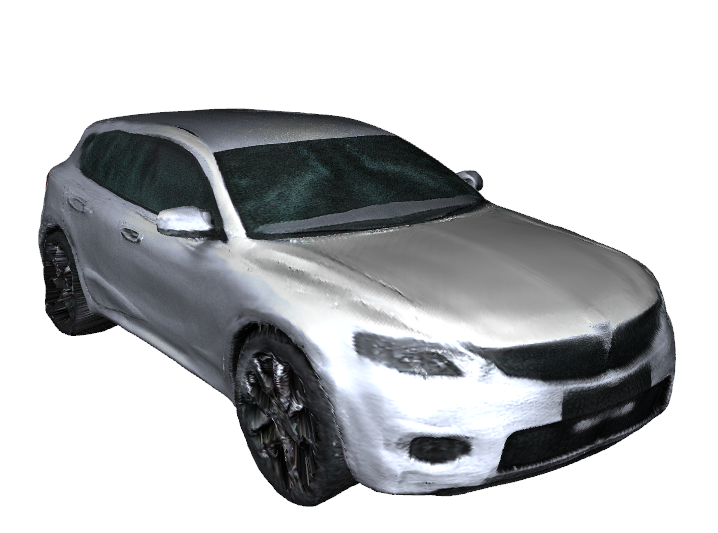}                    \\
      original                                                                 & projection (all) \\
      \includegraphics[width=0.22\textwidth]{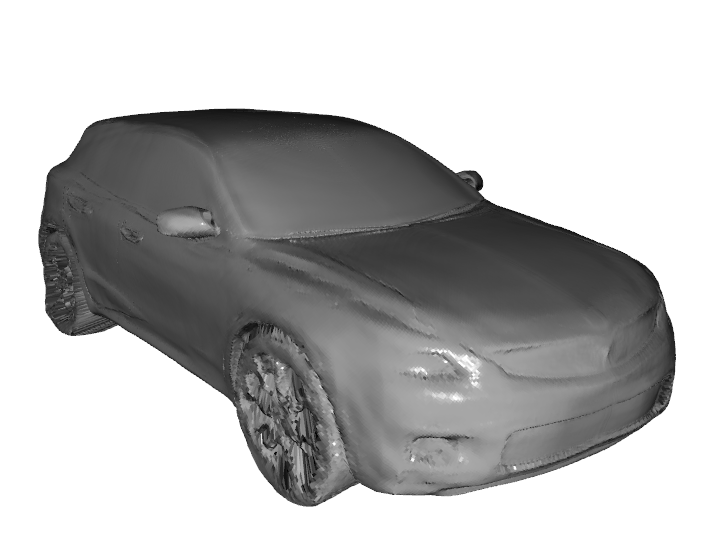}  &
      \includegraphics[width=0.22\textwidth]{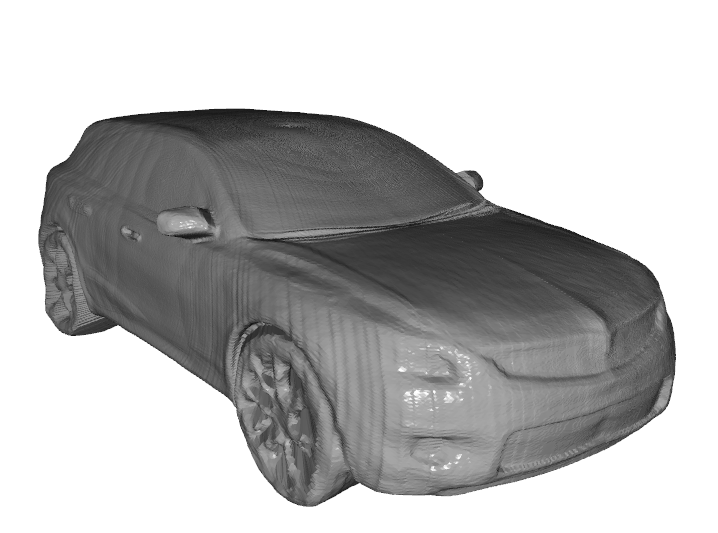}                 \\
      w/o texture                                                              & w/o normals      \\ 
    \end{tabular}
  }
  \vspace{0.8em}
  \caption{\ourmodel{} allows generation of high-resolution textured 3D meshes, supports projection of 3D models into the latent space, where generation of multiple textures and semantic editing are possible. Results from the model trained with HQCars dataset.}
  \vspace{-2.0em}
  \label{fig:teaser}
\end{figure}

We present \ourmodel{} (X-Dimensional GAN), a method for using standard fast 2D GAN architectures for generating high-resolution 3D meshes with additional surface properties like textures and normals. \ourmodel{} is trained on a collection of 3D shapes, which are first unwrapped into planar geometry images. Each pixel of a geometry image represents a vertex position of a fixed topology mesh, and is thus a direct representation of surface geometry. To our knowledge, ours is the first work to show that 2D convolutional generators can produce high-fidelity 3D shapes by operating on planar parameterization of 3D geometry, an idea that can help bring more advances in 2D architectures to the 3D realm. Specifically, we experiment with StyleGAN \cite{karras2019style,Karras2020AnalyzingAI} modified to train on higher-precision geometry images that represent 3D shapes. In order to augment generated geometry with additional per-vertex properties, we show that it is possible to directly apply an image-to-image translation framework \cite{park2019SPADE} to predict these properties for each geometry image. We demonstrate that our approach can generate diverse, high-quality textured 3D meshes, beating existing generators in quality of surface detail, presence of color or practical usefulness of output representation. Further, we demonstrate that powerful properties of the StyleGAN latent space can also be exploited for 3D generation, reconstruction, as well as supervised and unsupervised manipulation, just as it has been shown for 2D images.

\section{Related Work}\label{sec:related}

Generative Adversarial Networks (GANs) have become a popular technique for image generation since their introduction by Goodfellow et al.\ \cite{goodfellow2014gan}. 
Striking progress in the quality and resolution of GAN-generated images has been achieved in recent years \cite{BigGAN,karras2018progressive,karras2019style,Karras2020AnalyzingAI}, including in conditional settings \cite{park2019SPADE}. While some degree of view control of such models has been exploited for downstream 3D tasks \cite{zhang2020image}, these architectures have remained primarily confined to generation in the 2D domain. Most recently, approaches that leverage an intermediate 3D representation to improve 3D consistency and view control of GANs have also been proposed \cite{niemeyer2021giraffe,chan2021pi,chan2021efficient,gu2021stylenerf, alhaija2020intrinsic}. 
Our method is orthogonal to this line of work as we show that it is possible to leverage unmodified 2D GAN architectures, to learn \emph{3D geometry} directly. 
Beyond unconditional generation, latent spaces of large-scale GANs have been successfully used for both coarse and fine-grained manipulations of images. While some techniques focus on editing real images \cite{richardson2021encoding,kim2021exploiting,alaluf2021restyle,abdal2020image2stylegan++,ling2021editgan}, others devise meaningful exploration of the GAN latent space. Supervised approaches typically use pre-trained attribute classifiers to optimize edit directions \cite{goetschalckx2019ganalyze,shen2020interfacegan,shen2020interpreting}. 
Other works show that it is possible to find meaningful directions in latent space in an unsupervised way \cite{voynov2020unsupervised,jahanian2019steerability, harkonen2020ganspace}. In our work, we use a modified version of InterFaceGAN \cite{shen2020interfacegan} to demonstrate how latent space manipulation techniques, originally proposed for 2D image manipulation, can be leveraged for 3D generation as well.

While there have been remarkable advances in 3D shape generation in the recent years, deployment of 3D generative models poses challenges due to generation quality and speed, as well as compatibility of the output format with downstream tasks (see Tb.\ref{tb:methods} for summery).

\begin{table}[t!] 
\centering
\begingroup

\newcommand{\yes}{\color{green}\ding{51}}%
\newcommand{\no}{\color{red}\ding{55}}%
\newcommand{\methodrow}{ & }%

\small{
\setlength{\tabcolsep}{2pt} %
\begin{tabular}{@{}|l|c|c|c|cccc|@{}}
\Xhline{2.0pt}
 & \multicolumn{7}{c|}{\textbf{Representations} } \\ 
 \cline{2-8} 
 & Voxels &  Points  & Implicit & \multicolumn{4}{c|}{Mesh}\\
 \cline{2-8} 
& \cite{tatarchenko2017octree,zhang2021learning}
& \cite{Luo2021DiffusionPM}
& \cite{Park2019DeepSDFLC,Hao2020DualSDFSS,Chen2019LearningIF}
& \cite{sinha2016deep} & \cite{umetani2017exploring} &\cite{groueix2018} & \textbf{Ours}\\
\Xhline{2pt}
Real-time generation   & \yes    & \yes    & \no & \yes & \yes & \yes & \yes \\
Real-time rendering    & \yes & \yes & \no & \yes & \yes & \yes &  \yes \\
High-quality surface    & \no & \no & \yes & \no &  \no & \no &  \yes \\
Texture & \no & \no & \no & \no & \no & \yes  & \yes \\
Variable topology & \yes & \yes & \yes & \no & \no & \yes & \no \\
\Xhline{2pt}
\end{tabular}
}
\endgroup
\vspace{0.8em}
\caption{Comparison of 3D generative methods by the representation used. We include quantitative and qualitative comparisons with the starred methods in our experiments. }
\label{tb:methods}
\vspace{-1.2em}
\end{table}

The most direct 3D analog of the 2D pixel array is the 3D voxel grid, and early generative approaches adapted convolutional architectures to generate objects by operating on this 3D grid \cite{Wu20153DSA,Wu2016LearningAP}. Due to fixed resolution of the voxel grid and high memory demands of 3D convolutions these approaches are limited to low quality outputs. Later methods mitigate high memory requirements, for example, by using octrees to represent the 3D space more efficiently \cite{tatarchenko2017octree} or through local generation \cite{zhang2021learning}, but are still constrained in their ability to represent high-resolution smoothly varying geometry and texture (Tb.\ref{tb:methods}, A).

Due to their simplicity, point clouds, or unordered $(x, y, z)$ samples of surface geometry, are a popular representation for 3D generation \cite{Achlioptas2018LearningRA,cai2020learning,Li2021SPGANS3}, reconstruction \cite{Lin2018LearningEP} and segmentation \cite{fan2017point,qi2017pointnet,Qi2017PointNetDH}. Despite continued progress in generative modeling of point clouds \cite{pointflow,zhou20213d}, they cannot overcome the inherent limitations of the representation, which cannot model sharp surface details, or be rendered as high-fidelity textured meshes. We quantitatively compare our method against the recently published Diffusion Point Cloud (DPC) \cite{Luo2021DiffusionPM} (See also Tb.\ref{tb:methods}, B). DPC proposes a probabilistic diffusion model and uses point cloud representation of 3D shapes for shape generation, auto-encoding and shape completion.

Recently there has also been a surge in using implicit 3D representations, such as learned signed distance functions (SDFs) \cite{Park2019DeepSDFLC,Hao2020DualSDFSS} or occupancy  \cite{Chen2019LearningIF,Mescheder2019OccupancyNL}. Like point clouds and voxel grids, these representations allow modeling varying object topology, but can also represent high-resolution surface detail. Most related to ours among implicit methods is IM-GAN \cite{Chen2019LearningIF}, which proposes to use an implicit decoder IM-NET in conjunction with features learned by a latent-GAN model \cite{Achlioptas2018LearningRA} to yield a general 3D generative model. Because the outputs of this and related methods are implicit in the network weights, high-resolution results typically cannot be visualized or exported in real-time, requiring several seconds on state-of-the-art GPU, and precluding interactive applications (See Tb.\ref{tb:methods}, C). In addition, a number of hybrid approaches are emerging. For example, ShapeFormer \cite{yan2022shapeformer} employs transformers in the space of discrete implicit shape elements to support shape completion, and DMTet \cite{shen2021deep} produces detailed tetrahedral meshes using sparse SDF samples. While DMTet generates a mesh-based representation quickly, like our method, it is only designed to work when conditioned on a rough input shape, not as a general generator.

\vspace{-1.0em}
\section{Methodology}\label{sec:method}

\begin{figure*}[t]
\centering
\subfloat[Training \ourmodel{}.]{%
  \includegraphics[width=0.95\linewidth]{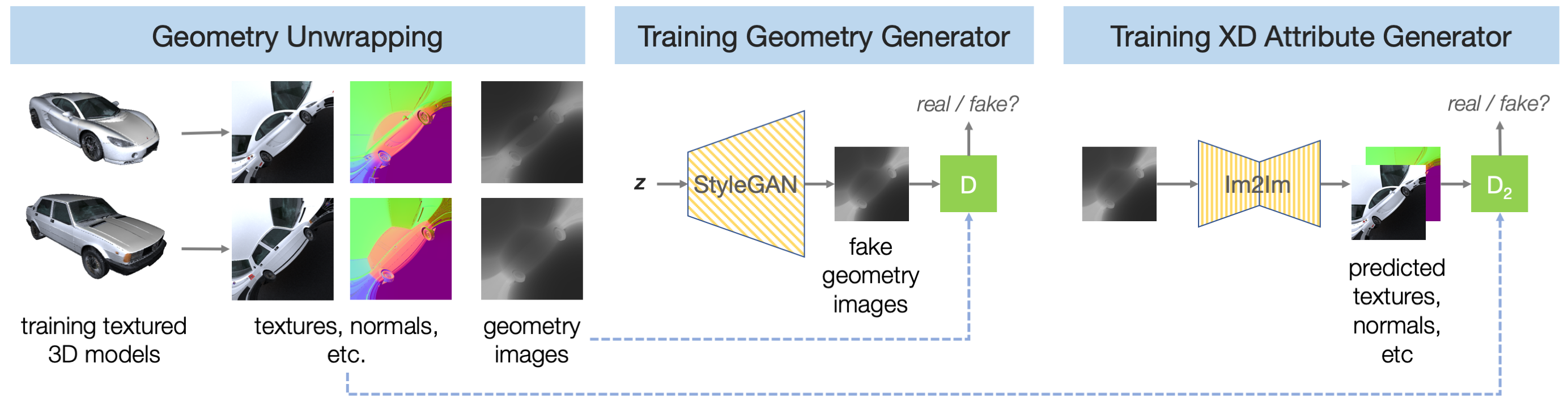}%
  }\\
  \vspace{-0.9em}
  \subfloat[Test-time generation using \ourmodel{}.]{%
  \includegraphics[width=0.95\linewidth]{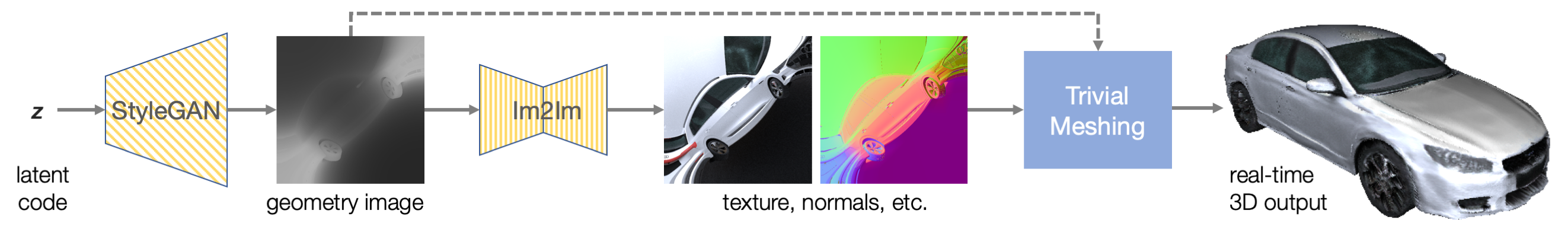}%
  }
\vspace{0.8em}
\caption{\textbf{Overview of \ourmodel{}:} To train our model (a), we first convert a training dataset of textured 3D meshes into 2D geometry images with corresponding textures, normals or any other surface attributes. Next, we train a GAN model on geometry images to generate geometry, and an image-to-image translation network to generate X-channel attribute images for an input geometry image. At test-time (b), feed-forward evaluation of these networks followed by a trivial meshing step produces textured 3D output meshes in real-time.}
\label{fig:arch}
\vspace{-1.8em}
\end{figure*}

Our approach leverages the power of 2D convolutional generator networks in order to generate 3D geometry. Given a training set of (optionally textured) 3D objects, we first convert them into geometry images and corresponding normal maps, textures or other attribute maps (\S\ref{ssec:geo-img}). The one-channel geometry images represent vertex positions of a fixed topology mesh, and are used to train an image-based GAN model to generate such plausible geometry images (\S\ref{ssec:stylegan}). In order to augment output with per-vertex attributes like textures and normals, we similarly train an image-to-image translation network on these aligned attribute maps (\S\ref{ssec:stylegan}). After training (Fig.\ref{fig:arch}a), these two image-based generators produce detailed textured 3D meshes in real-time (\S\ref{fig:arch}b). Fast generation makes interactive exploration of the GAN latent space especially attractive. Although our method generates 3D output, we reap the benefits of latent space exploration techniques developed for the 2D domain to explore and edit 3D shapes (\S\ref{ssec:latent}).

\vspace{-0.8em}
\subsection{Conversion to Geometry Images}\label{ssec:geo-img}

Geometry images \cite{Gu2002GeometryI} are a representation that captures 3D surface as a 2D array of $(x, y, z)$ values via an implicit surface parameterization, which is a bi-directional mapping from an object's surface to a 2D plane. Such parameterizations are often used in computer graphics to map 2D textures and other attributes onto a 3D surface mesh. Once the parameterization is determined, the plane is sampled to create a 3-channel $n \times n$ image where each pixel represents an $(x, y, z)$ vertex location of a new geometry image mesh, which is independent from the original object's mesh topology or representation. 
The order and connectivity of each vertex within this geometry image mesh is encoded by its pixel index, where faces connect neighboring pixel-vertices using a regular structure. Geometry images can contain additional channels encoding surface properties like color texture, normals, object segmentation and similar. Because the same surface parameterization is used when sampling these additional properties, all channels of the geometry image are aligned, with each pixel representing location and other properties of the same 3D vertex. In our work, we generate normals and texture geometry images and propose applying 2D convolutional generators to these aligned X-channel images.

\begin{wrapfigure}{R}{0.5\textwidth}
\centering
  \includegraphics[width=\linewidth]{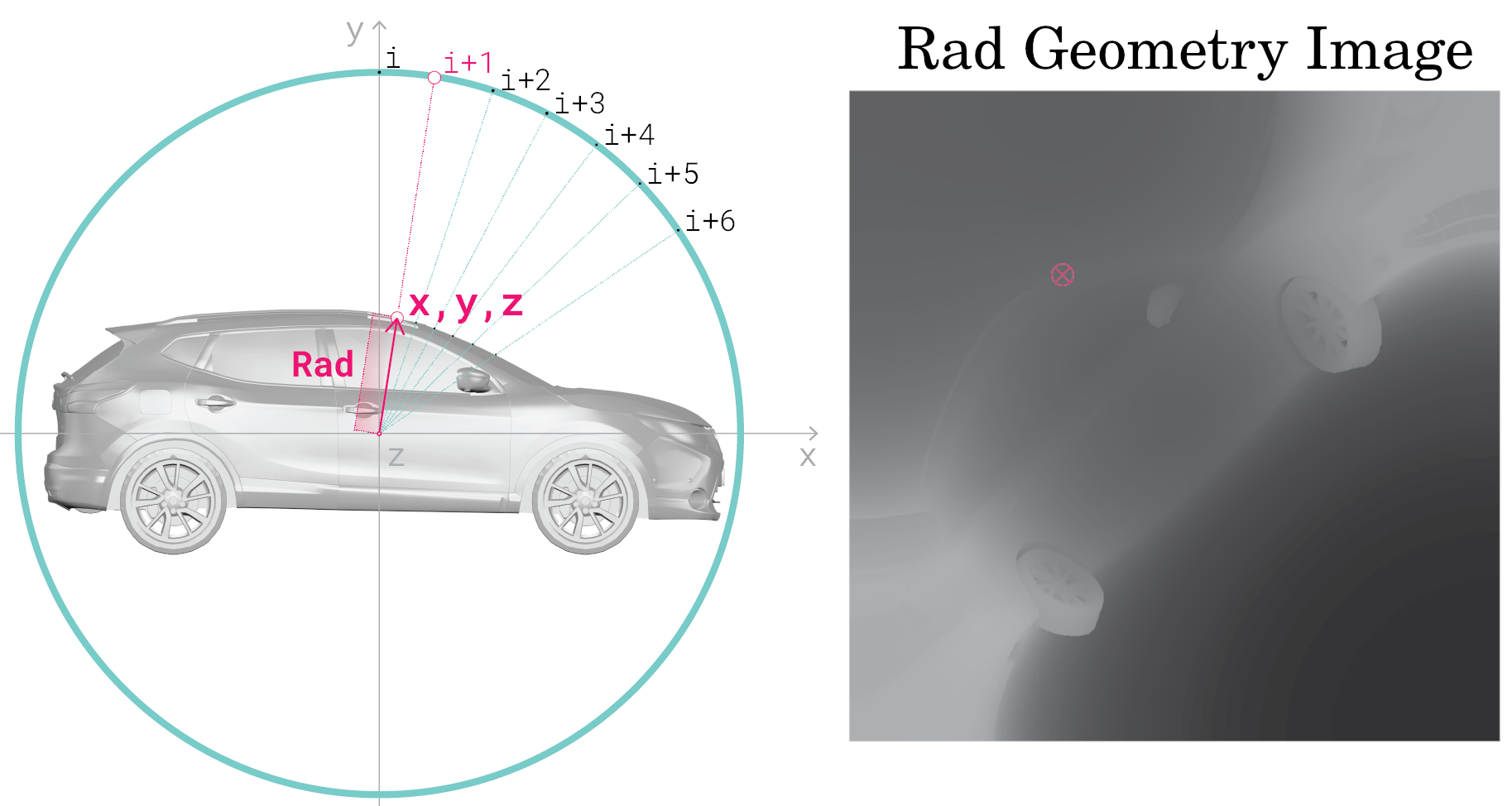}%
\caption{Generation of geometry images from 3D data. See \S\ref{ssec:geo-img}.}
\label{fig:geo_img}
\vspace{-0.8em}
\end{wrapfigure}

In order to map a closed 3D surface onto a 2D plane, we need to determine seams, make cuts across the surface to conform to a chosen topology (e.g.\ disk, sphere, cylinder), and unwrap it onto a flat square surface. Creating appropriate parameterizations that are well-suited to individual 3D object topology and shape is an active research area \cite{Gu2002GeometryI,SorkineHornung2002BoundeddistortionPM,Sheffer2002SeamsterIL,Poranne2017AutocutsSD,Li2018OptCutsJO}. Since our goal is to train a generative model, we want to reduce the representation variability as much as possible to allow the model to learn the true data distribution instead of the variability in parameterizations. For this reason, we choose to work with a specific family of spherical parameterizations, and a fixed seam structure to convert 3D shapes to geometry images and create consistent input for training the 2D generator. We start with two realistic assumptions about the geometry of our shapes. First, we assume that most of our 3D shapes are symmetric, which means we only need to represent half of the shape in the geometry image. Cutting the closed surface of the shape in half also defines an edge of the surface that is needed for the parameterization. Second, we assume our shapes are mostly genus-zero or close to it. This means that the surface can be mapped onto a surface of a sphere.  While the spherical projection method imposes limitations on output geometry, our aim is to explore generation in the most consistent setting first. 

Given above assumptions, we define a three step process to obtain a geometry image from a collection of 3D shapes (Refer to Fig.\ref{fig:geo_img}). 
We first scale and align all input shapes to ensure consistent orientation and scale and place them within the projection sphere. We then project the surface of the shape onto a spherical domain using the normal direction of each point on the sphere. This creates a unique mapping between each point on the sphere's surface and the shape's surface. Finally, we cut the sphere in half along the symmetry axis of the shape and map the surface of the hemisphere onto a 2D plane using a fixed, predefined area-preserving parameterization. Because projection from the sphere onto a 3D surface has only one degree of freedom (distance along the normal direction of the sphere, see Fig.\ref{fig:geo_img}), we represent the 3D location of a surface point as a single value measuring the distance (which we call ``Rad") from the origin along the projection ray, effectively reducing the degrees of freedom of the representation from 3 channels per pixel to only 1. To compute the original $(x,y,z)$ coordinates of all the vertices, one only needs to multiply the Rad image by the image representing the normals of the projection sphere, fixed for the whole dataset. In addition to positions, we produce aligned images of normals and color textures.

\vspace{-0.8em}
\subsection{Generating Geometry Images and Other Attributes}\label{ssec:stylegan}

Once our training dataset is converted to Rad geometry images (\S\ref{ssec:geo-img}), we train standard 2D GAN architectures to produce "realistic" Rad images. The only modification required is to adapt them to high-precision geometry images instead of the traditional 8-bit RGB images. Specifically, we experiment using StyleGAN3 \cite{karras2021alias} for geometry generation, but nothing suggests that our approach will not generalize to other GAN architectures (see Supplemental for experiments with StyleGAN2 \cite{Karras2020AnalyzingAI}).
We found training over single-channel Rad images to differ little in its stability from standard image-based GAN training.

During the generation of Rad geometry images (\S\ref{ssec:geo-img}), we also produce aligned normal and texture maps using the same surface parameterization. Normal maps describe the local curvature of a 3D surface and are often used in computer graphics to add fine details to a 3D shape without adding actual geometry. To generate smooth normal maps even for shapes with low resolution, we use the smooth shading technique which interpolates the normals along a flat mesh triangle to give the illusion of smooth shading. The computed smooth normals are then mapped onto a 2D image using the same parameterization. The texture maps are generated by using an environment map around the object which can generate both realistic lighting and reflection for the object. The result of the rendering is then baked as a texture and mapped into a 2D image using the same procedure as with normal maps.

We use these aligned image channels to train an image-to-image translation network to produce X attribute channel images given an input Rad image of the same resolution. Specifically, we experiment with SPADE image-to-image translation network \cite{park2019SPADE}, but nothing in our setup is specific to this particular network choice. We note that SPADE is designed to be able to produce diverse outputs given a single input.

\vspace{-0.8em}
\subsection{Exploiting the Latent Space}\label{ssec:latent}

A major advantage of our 3D generative model is that it is based on standard GAN architectures, which allows us to directly apply various techniques developed for image-based GANs.  To demonstrate this, we integrate the existing technique of InterFaceGAN \cite{shen2020interfacegan} in order to find editing directions within our trained latent space. To this end, we first project our training shapes into the latent space $W$ to create a set of of projected latent vectors $w$ with lables (e.g.\ \textit{sports-car, SUV, \etc }). We then train one linear SVMs for each label using these labeled examples, and use the unit normal vector to the separation hyperplane of the SVM $\mathbf{n}^{T}$ as an editing direction in latent space. The target attribute of a real or generated 3D shape can then be enhanced or negated by simply steering its latent vector $p$ along this editing direction $\mathbf{n}^{T}$ by $t$ steps with a step size of $\alpha$ to produce a new latent vector
    $p^{'} = p + (\alpha \times t \times \mathbf{n}^{T})$

\section{Experiments}
We evaluate our model's performance on shape projection, editing and single-view reconstruction tasks. 
We compare our method to three state-of-the-art 3D shape generation methods that use triangular meshs (AtlasNet \cite{groueix2018}), point-clouds (DPC \cite{Luo2021DiffusionPM}) and implicit representation (IM-Net \cite{Chen2019LearningIF}). The experiments show that our method achieves significantly better results on some object classes and competitive results on others, all while performing inference several orders of magnitude faster than the closest competitor IM-NET \cite{Chen2019LearningIF}. 
\vspace{-1.0em}
\subsection{Datasets and Experimental Setup}\label{sec:exp:datasets}

For the quantitative comparisons, we use the ShapeNetV2 \cite{chang2015shapenet} \textit{car} and \textit{airplane} categories consisting of 3509 and 4045 shapes respectively and use 10\% of each set for testing. We choose these two categories as they are sufficiently distinct structurally and demonstrate the range of our method. Additionally, we use a proprietary new dataset consisting of 3050 high-quality car models which we call \textit{HQCars}. In addition to the higher quality of meshes, this dataset provides high quality texture maps and semantic labels, like model, maker, year and type, which are not available in ShapeNet.

To create our training set, we first converted the set of 3D meshes to 1-channel Rad geometry images of size $512$ x $512$ and the corresponding normals and texture maps. (See Supplementary Material for comparisons with geometry images using 3 $(x,y,z)$ channels)
We use the Rad geometry images to train a modified version of the official StyleGAN3 implementation \cite{karras2021alias} to process high-precision values instead of the 8-bit RGB images. We train this StyleGAN3-r variant without any data augmentation with batch size 8 and gamma value of 1 for 500 epochs.
To generate the additional texture and normal maps, we  train an image-to-image translation model separately for each using the official implementation of SPADE \cite{park2019SPADE}, similarly modified to deal with the range and precision of geometry images. Each model is trained on aligned pairs of ground-truth geometry images and texture or normals maps, for 300 epochs with batch size of 32.

We train our model and all baselines from scratch on the \textit{car} and \textit{airplane} training data subsets \textit{separately} and evaluate on the respective test subsets. We use the official Pytorch implementations of AtlasNet, DPC and IM-Net with the default parameters.

For all experiments, we use reconstruction-based tasks and compare the generated shapes to ground truth shapes using Chamfer distance (CD) and Earth Mover's Distance (EMD) metrics. CD and EMD metrics are computed by sampling 5000 points from the generated and ground truth shapes, matching the nearest points across the sets of points and computing their sum of Euclidean and Manhattan distances respectively. In addition to CD and EMD, we use the Light Field Distance (LFD) \cite{Chen2003OnVS} as a visual similarity metric. To compute LFD, the generated and ground truth shapes are rendered from various camera angles. The rendered images are then encoded using Zernike moments and Fourier descriptors to compute similarity. Finally, we report the average inference time of ours and competing methods as real-time generation is critical for interactive applications.
\subsection{Quantitative Evaluation}\label{ssec:quant_eval}
For evaluating on the task of 3D shape reconstruction, we use an iterative latent projection approach to obtain a reconstruction of ground-truth test model. We start from a randomly sampled latent vector and iteratively compute the loss between the generated geometry image from this vector and the ground-truth geometry image, then back-propagate the loss through the generator to update the latent vector. We repeat this procedure for 500 iterations for each test model. For the single-view reconstruction task, we train a ResNet-based encoder on the rendered RGB images \cite{choy20163d} of a ShapeNet model from a random viewpoint, and predicts a latent vector which is then used to generate a geometry image. For both tasks, the final evaluation metrics are computed between the triangle mesh created from the generated geometry image and the original ShapeNet test mesh.

\begin{table}[t]
\small{
\centering
\begingroup
\setlength{\tabcolsep}{6pt} %
\renewcommand{\arraystretch}{1.2}

\begin{tabular}{@{}|l|c|c|c|c|c|c|c|@{}}
\Xhline{2.0pt}
 & \multicolumn{6}{c|}{\textbf{Reconstruction error} } & \textbf{Infr. Time}\\
 \cline{2-8} 
 & \multicolumn{2}{c|}{Chamfer $\downarrow$} & \multicolumn{2}{c|}{EM Dist. $\downarrow$} & \multicolumn{2}{c|}{LFD Dist. $\downarrow$} & Seconds $\downarrow$ \\
 \cline{2-8} 
            & Car       & Airplane      & Car       & Airplane & Car        & Airplane   & Mean \\
\Xhline{2.0pt}
AtlasNet-Sph & 2.089     & 6.241      & 2.43      & 4.85       & 2610       & 5006       & 0.078  \\
AtlasNet-25 & 1.786      & 5.216      & 2.22      & 3.94       & 2565       & \bf{4730}  & 0.112  \\
IM-Net      & 1.439      & \bf{0.406} & 1.49      & \bf{1.06}  & \bf{2243}  & 4759       & 13.280  \\
DPC         & 6.146      & 4.450      & 4.58      & 4.11       & -          & -          & 0.061  \\
Ours        & \bf{0.977} & 2.721      & \bf{1.34} & 1.96       & 2407       & 7435       & \bf{0.020}  \\
\Xhline{1pt}
\end{tabular}
\endgroup
\newline
\caption{Quantitative results on ShapeNet reconstruction. Note that the CD values are multiplied by $10^{3}$ and EMD are multiplied by $10^{2}$. Since DPC produces only point clouds, it is not possible to compute the LFD distance for its outputs. }
\label{tb:quant}
}
\vspace{-2em}
\end{table}

Results in Table \ref{tb:quant} show that our method achieves by far the best CD and EMD on the car category and ranks second on the airplane category after IM-Net, while being faster by factor of 500. This finding is not surprising, as we have made some simplifying assumptions during conversion to geometry images (\S\ref{ssec:geo-img}), and it is harder to map airplanes with many sharp and spiky features onto the spherical domain than it is for relatively compact cars. These results also demonstrate that when the 2D parameterization is appropriate (as it is for cars), 2D generative models using our technique outperform even implicit 3D representation models, which is an unexpected outcome. We further note that our method is the fastest by at least a factor of 3 over any baseline. For Single-View reconstruction experiment (Table \ref{tb:quant-svr}), the results are similar where our methods achieves very good results on the car category while falling short from IM-Net on the airplane category.
\begin{table}[h]
\centering
\begingroup
\setlength{\tabcolsep}{15pt} %
\renewcommand{\arraystretch}{1.2}

\begin{tabular}{@{}|l|c|c|c|c|@{}}
\Xhline{2.0pt}
 & \multicolumn{4}{c|}{\textbf{SVR error} } \\
 \cline{2-5}
 & \multicolumn{2}{c|}{Chamfer $\downarrow$} 
 & \multicolumn{2}{c|}{EM Dist. $\downarrow$}  \\
 \cline{2-5} 
                & Car       & Airplane  & Car       & Airplane  \\
\Xhline{2.0pt}
AtlasNet-Sph    & 2.335     & 5.514     & 2.558     & 4.917  \\
AtlasNet-25     & 2.107     & 4.605     & 2.264     & 3.737  \\
IM-Net          & 1.437     & \bf{0.907}&\bf{1.492} & \bf{1.282}  \\
Ours            & \bf{1.325}& 3.055     & 1.511     & 2.030  \\
\Xhline{1pt}
\end{tabular}
\endgroup
\newline
\caption{Quantitative results on Single View Reconstruction. Note that the CD values are multiplied by $10^{3}$ and EMD are multiplied by $10^{2}$.}
\label{tb:quant-svr}
\vspace{-2em}
\end{table}

\subsection{Qualitative Results}\label{ssec:qual_eval}

Table \ref{tb:qual_proj} show 3D meshes projected with XDGAN using the method in \S\ref{ssec:quant_eval}, visually compared to other competitors. 
Our methods generates water-tight meshes with high-fidelity that preserve even small details like car mirrors and wheel patterns. The inclusion of generated surface normals also add another level of detail and smoothness where needed, making the results from XDGAN visually outstanding compared to other methods.

\begin{figure}[t]
\centering

  \includegraphics[width=0.82\textwidth]{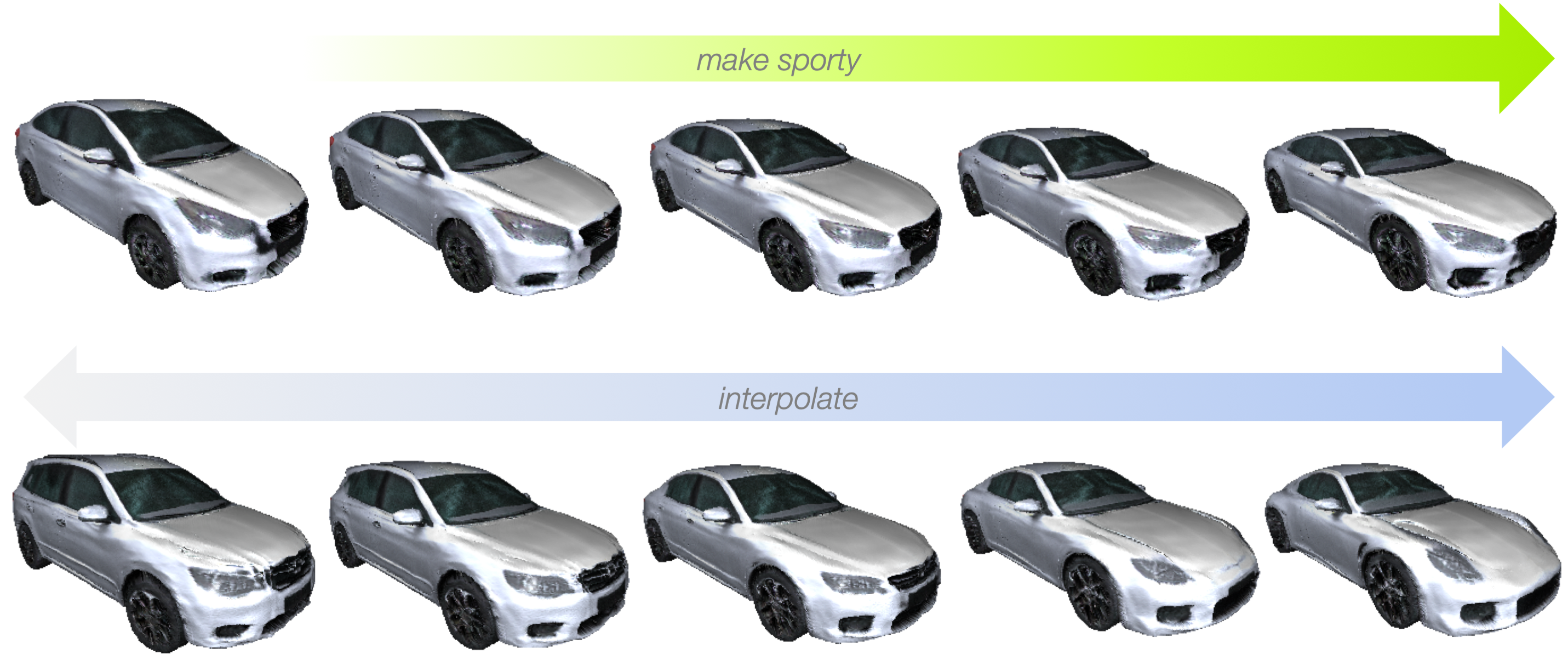}
\caption{The latent space of \ourmodel{} allows meaningful latent space exploration such as semantic editing (top, see \S\ref{ssec:latent}) and interpolation between two input models (bottom).}
\label{fig:editing}
\end{figure}

Furthermore, Fig.\ref{fig:editing} shows the the powerful properties of \ourmodel{}'s latent space. Editing directions optimized using InterFaceGAN \cite{shen2020interfacegan} (\S\ref{ssec:latent}) allow semantic editing of the 3D object \emph{in real time}, for example by making a car appear sportier (Fig.\ref{fig:editing}, top). 
The latent space additionally allows for generating intermediates between two input models by linearly interpolating their projected latent space vectors.

\begin{table*}[t]
  \centering
  \newcommand{\qualimg}[1]{\raisebox{-.5\height}{\includegraphics[width=0.13\linewidth]{#1}}}
  \begin{tabular}{m{2cm}cccccc}
  Input Model &
  \qualimg{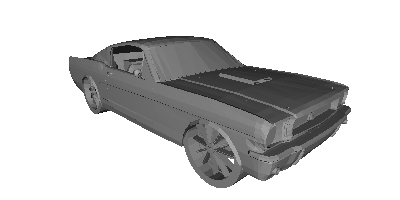} & 
  \qualimg{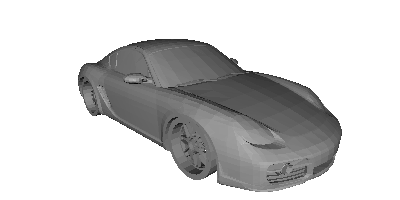} & 
  \qualimg{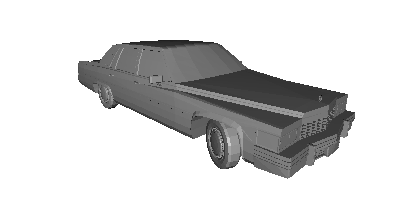} &  
  \qualimg{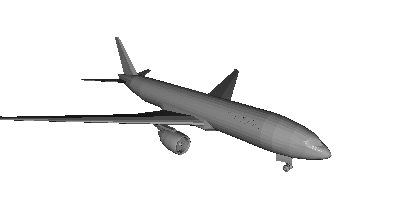} & 
  \qualimg{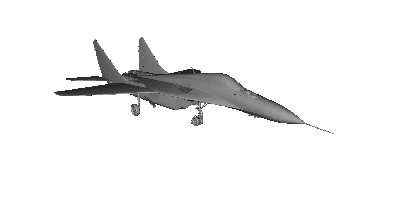} \\
  \hline
  AtlasNet-Sphere &
  \qualimg{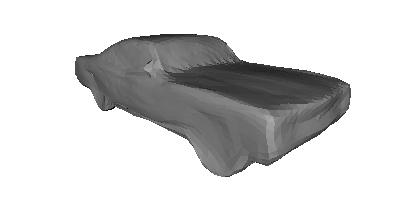} & 
  \qualimg{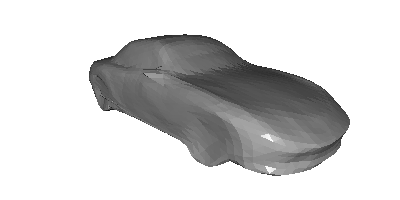} & 
  \qualimg{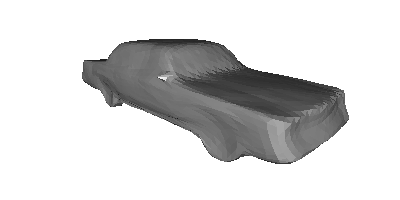} & 
  \qualimg{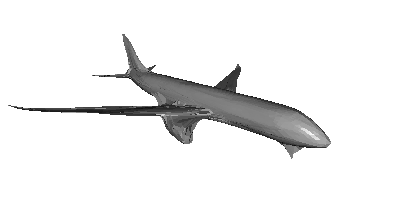} & 
  \qualimg{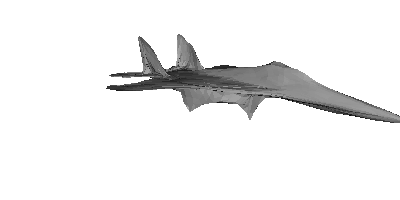} \\
  AtlasNet-25 &
  \qualimg{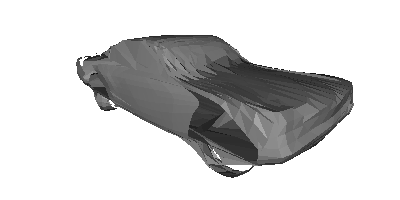} & 
  \qualimg{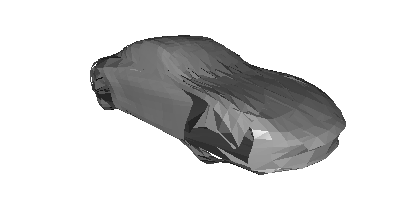} & 
  \qualimg{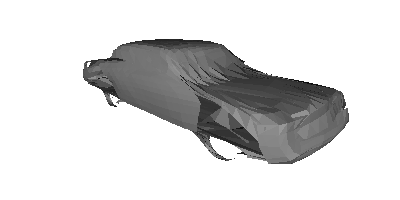} & 
  \qualimg{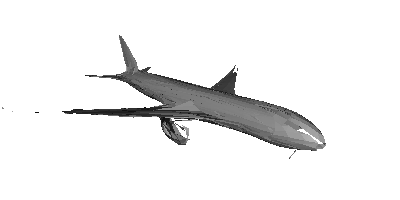} & 
  \qualimg{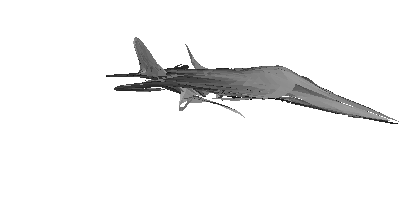} \\
  IM-Net &
  \qualimg{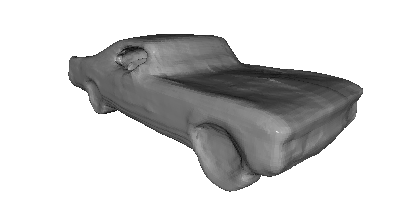} & 
  \qualimg{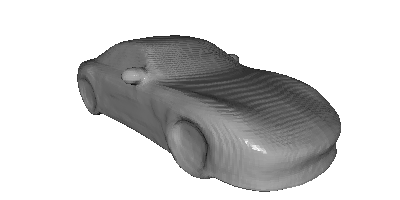} & 
  \qualimg{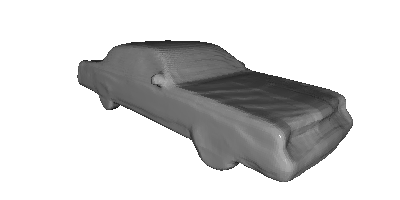} & 
  \qualimg{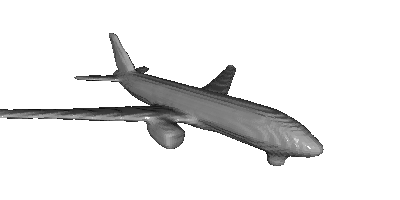} &
  \qualimg{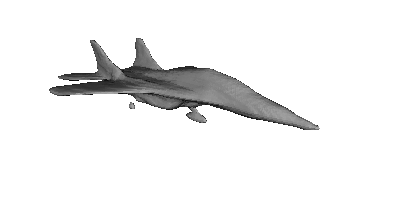}   \\
  \hline
  \makecell[l]{\begin{tabular}{@{}l@{}}Ours \\(no normals)\\ \end{tabular}} &
  \qualimg{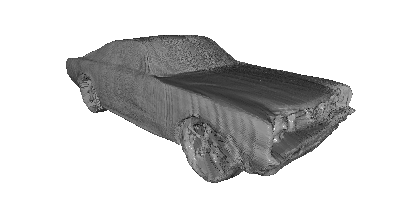} & 
  \qualimg{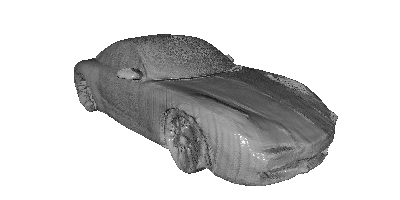} & 
  \qualimg{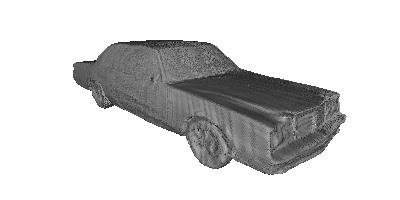} &  
  \qualimg{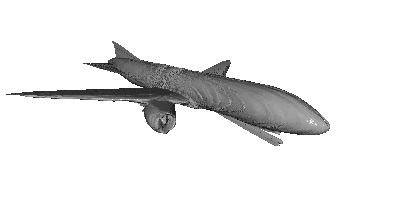} & 
  \qualimg{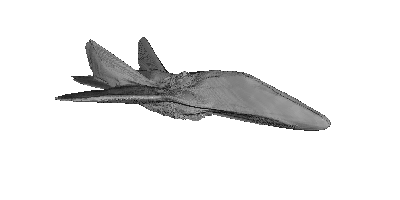}  \\
  Ours & 
  \qualimg{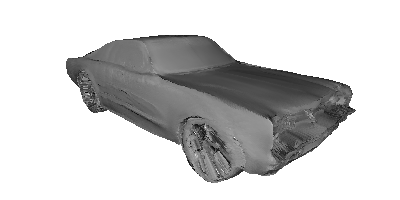} & 
  \qualimg{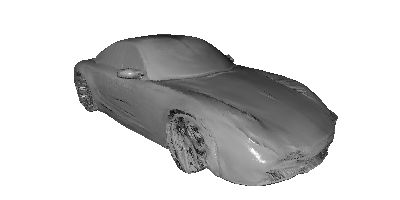} & 
  \qualimg{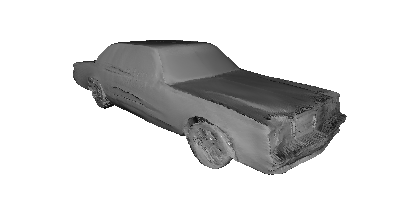} &  
  \qualimg{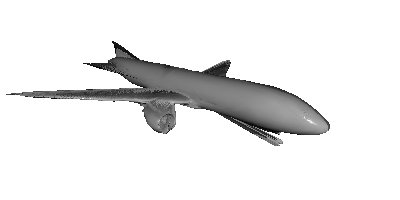} &
  \qualimg{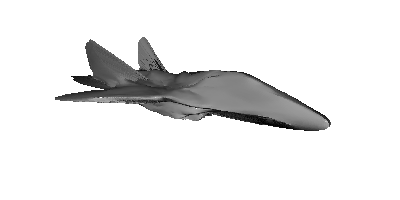}  \\
  \end{tabular}
  \vspace{0.2em}
  \caption{Qualitative Comparison of Reconstruction Results (Zoom for more details). Note that our method can generate highest-fidelity 3D meshes including detailed surface normals.}
  \label{tb:qual_proj}
 \vspace{-1em}
\end{table*}

\section{Conclusion}

We presented \ourmodel{}, the first method to show that 2D convolutional architectures can be directly used for generative modeling of 3D objects with high fidelity. Our approach is based on a novel combination of planar image parameterization with an image-based GAN. We show that our method generates high-quality textured 3D meshes, allows unconditional sampling, projection, and can be used in several conditional settings. In addition, \ourmodel{}'s speed and convenient mesh output allow for a number of applications that can be directly employed in practice.
The shortcoming of our approach is its reliance on a fixed topology mesh, associated with the geometry image. We believe that there are many ways future work could mitigate this limitation, for example through part-based generation. Despite this limitation, we have shown impressive generation results from our method, and hope that this work may help close the gap between advances in 2D generative architectures and comparable approaches in 3D. 

\section{acknowledgments}

We thank Stefan Liske and Timotheus Gmeiner from PCH Innovations GmbH for the insightful discussions and making this collaboration possible. Part of this research work has been done while Hassan Abu Alhaija was at PCH Innovations GmbH.
\bibliography{bibliography}
\end{document}